\documentclass[10pt,twocolumn,letterpaper]{article}

\usepackage{authblk}

\usepackage[pagenumbers]{cvpr} 
\makeatletter
\@namedef{ver@everyshi.sty}{}
\makeatother
\usepackage[accsupp]{axessibility}  
\usepackage[T1]{fontenc}
\usepackage{graphicx}
\usepackage{amsmath}
\usepackage{amssymb}
\usepackage{booktabs}
\usepackage[acronym]{glossaries}

\usepackage[pagebackref,breaklinks,colorlinks]{hyperref}

\usepackage[capitalize]{cleveref}
\crefname{section}{Sec.}{Secs.}
\Crefname{section}{Section}{Sections}
\Crefname{table}{Table}{Tables}
\crefname{table}{Tab.}{Tabs.}

\def\cvprPaperID{2} 
\def\confName{CVPM}
\def\confYear{2023}

\newacronym{ecg}{ECG}{electrocardiogram}
\newacronym{shhs}{SHHS}{Sleep Heart Health Study}
\newacronym{rip}{RIP}{Respiratory inductance plethysmography}

\begin{document}

\title{Deep Learning-Enabled Sleep Staging From Vital Signs and Activity Measured Using a Near-Infrared Video Camera}

\author[1,2]{Jonathan Carter\thanks{jcarter@robots.ox.ac.uk}}
\author[2]{Jo\~{a}o Jorge}
\author[2]{Bindia Venugopal}
\author[2]{Oliver Gibson}
\author[1]{Lionel Tarassenko}
\affil[1]{Institute of Biomedical Engineering, University of Oxford}
\affil[2]{Oxehealth Ltd., Oxford}

\maketitle
\begin{abstract}
   Conventional sleep monitoring is time-consuming, expensive and uncomfortable, requiring a large number of contact sensors to be attached to the patient. Video data is commonly recorded as part of a sleep laboratory assessment. If accurate sleep staging could be achieved solely from video, this would overcome many of the problems of traditional methods. In this work we use heart rate, breathing rate and activity measures, all derived from a near-infrared video camera, to perform sleep stage classification. We use a deep transfer learning approach to overcome data scarcity, by using an existing contact-sensor dataset to learn effective representations from the heart and breathing rate time series. Using a dataset of 50 healthy volunteers, we achieve an accuracy of 73.4\% and a Cohen's kappa of 0.61 in four-class sleep stage classification, establishing a new state-of-the-art for video-based sleep staging.
\end{abstract}

\section{Introduction}\label{section:intro}
The `gold-standard' method for sleep monitoring is video polysomnography (vPSG,~\cite{stefani_prospective_2015}), a test in which a number of physiological signals are recorded over a night's sleep. This typically requires an extensive set of contact sensors to be attached to the patient, which often leads to patient discomfort and an unrepresentative night's sleep \cite{agnew_first_1966}. After the recording, a human expert (sleep physiologist) must review the outputs from these sensors, manually annotating the record with sleep stages, leg movements and other events of interest from throughout the night. This process can often take multiple hours to complete. The cost of the study and the discomfort to the patient mean that longer-term studies are rare. Easier longitudinal monitoring would also benefit clinical research, from making trials for sleep-disorder treatments easier, to detecting the onset of dementia, by monitoring variations in slow-wave sleep~\cite{shinar_automatic_2001}. 

Sleep stages are commonly annotated at 30-second intervals using the American Academy of Sleep Medicine (AASM) guidelines. These divide sleep into five discrete states: Wake, N1, N2, N3 and REM sleep. N1 to N3 account for varying degrees of non-rapid eye movement (NREM) sleep from near-wakefulness (N1) to deep slow-wave sleep (N3).

There is well-studied subjectivity between sleep physiologists in scoring sleep stages, which is often reported using the Cohen's kappa statistic~($\kappa$,~\cite{cohen_coefficient_1960}). The work of Danker-Hopfe \etal~\cite{danker-hopfe_interrater_2009} showed that pairs of scorers agreed on 82\% of epoch-by-epoch sleep stages ($\kappa=0.76$) across a dataset of 72 sleep recordings.

A recent review has claimed that deep learning methods can score PSG records as accurately as a human scorer~\cite{phan_automatic_2022}. The SleepTransformer model~\cite{phan_sleeptransformer_2022} achieves a Cohen's $\kappa$ of 0.83 in AASM sleep stage classification on the Sleep Heart Health Study (SHHS,~\cite{quan_sleep_1997}) dataset from an electroencephalogram (EEG) signal. However, the EEG is typically measured using electrodes which must be glued to the subject's scalp by a sleep physiologist. 

To overcome this need, other work has investigated using cardio-respiratory and activity information obtained from wearable devices~\cite{walch_sleep_2019, pini_automated_2022, wulterkens_it_2021, radha_deep_2021}. Wearable sleep staging is possible from these inputs since they encode information about the underlying sleep state. For example, respiration tends to be more irregular during REM sleep~\cite{hudgel_mechanics_1984}, whilst low frequency power in the heart rate time series is typically lowest during N3 sleep~\cite{shinar_automatic_2001}. Wearable-based methods have commonly simplified sleep staging into a four-class classification problem, by merging N1 and N2 into a `light sleep' class. Doing so, Radha \etal~\cite{radha_deep_2021} achieve a Cohen's $\kappa$ of 0.65 in four-class sleep staging from wrist-worn photoplethysmography.\newline

\begin{figure*}[t]
  \centering
   \includegraphics[width=1.0\linewidth]{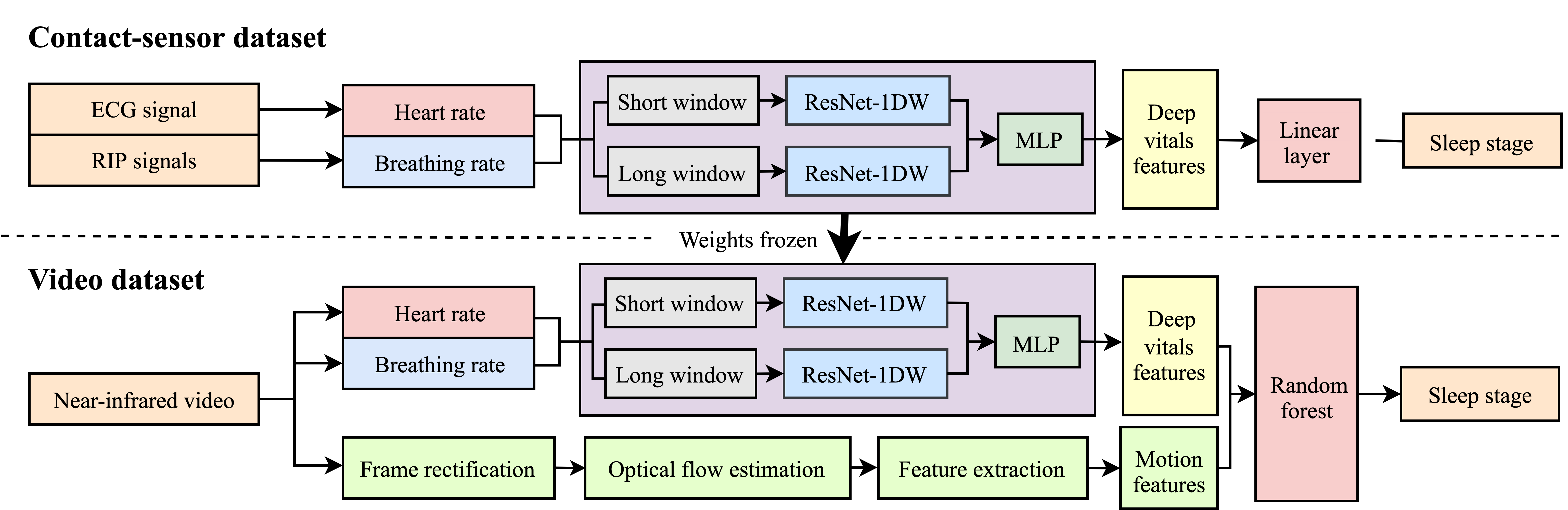}
   \caption{Overview of our approach. We first train a deep learning model on a contact sensor dataset to classify sleep stages from heart rate and breathing rate time series. We then use this trained model as a feature extractor to produce `deep' features from the video-derived vital signs of our target dataset. To classify sleep stages from video data, we train a random forest classifier which uses these deep features in conjunction with motion features derived from optical flow.}\label{fig:transferlearning}
\end{figure*}

Prior work has shown that both cardio-respiratory~\cite{van_gastel_camera-based_2021} and activity information~\cite{long_video-based_2019} can be derived from video data during sleep, indicating that video could be used as a single source from which to classify sleep stages. As an entirely non-contact solution, this would further simplify and improve the sleep monitoring process for both patients and clinicians. Additionally, video remains an important modality for the diagnosis of conditions including periodic limb movement disorder, parasomnias and REM behaviour disorder~\cite{stefani_sleep_2020, delrosso_video-polysomnographic_2019}. It may not be feasible to identify these conditions using alternative solutions such as wrist-worn wearables.

In this work, we use heart rate, breathing rate and activity signals, all derived from near-infrared video, to perform sleep stage classification. We use optical flow~\cite{horn_determining_1981} to derive continuous measures of activity during sleep, which are made robust to camera position and bed orientation using image homography transformations~\cite{hartley_multiple_2003}.

To overcome data scarcity, we adopt a transfer learning approach (Fig.\,\ref{fig:transferlearning}). We first train a deep neural network to classify sleep stages from heart rate and breathing rate time series derived from a substantially larger contact sensor dataset. The trained model is then applied to heart rate and breathing rate time series derived from near-infrared video to obtain `deep' features. These deep features are used in combination with video-derived activity features as inputs to a classifier which is trained and evaluated on a video dataset of 50 healthy volunteers using 10-fold cross validation.

Using our approach, we present improved results over the existing state-of-the-art~\cite{van_meulen_contactless_2023} for video-based four-class sleep staging in terms of both accuracy (73.4\% vs. 67.9\%) and Cohen's kappa (0.61 vs. 0.49).

\section{Related work}
\textbf{Video actigraphy:} 
Counting pixel differences above a threshold has proven to be a simple but effective activity signal for binary sleep-wake classification from both near-infrared (NIR,~\cite{schwichtenberg_pediatric_2018}) and depth images~\cite{kruger_sleep_2014, veauthier_contactless_2019}.

Optical flow~\cite{nakajima_development_2001, cuppens_automatic_2010} and spatio-temporal recursive search algorithms~\cite{heinrich_body_2013} have also proved effective at quantifying activity. These methods both compute motion vectors from the video feed and use average magnitudes as activity signals. The latter method has been used to detect periodic limb movements during sleep~\cite{heinrich_robust_2014}.

Using activity features derived from video, Long \etal~\cite{long_video-based_2019} achieve an accuracy of 92.0\% in sleep-wake detection from video using a dataset of 10 healthy infants.

\textbf{Video-based sleep staging:} To the best of our knowledge, only two prior works have attempted to classify sleep beyond a binary sleep-wake classification using a video camera. These works differ greatly in the information which they extract from the video to classify sleep stages.

Nochino \etal~\cite{nochino_sleep_2019} use frame-differencing to derive an activity signal from the camera. From this signal, the authors derive a number of motion features for each sleep epoch and train a Support Vector Machine to perform four-class sleep staging, achieving an average accuracy of 40.5\% and a Cohen's $\kappa$ of 0.19 on a dataset of 6 participants.

van Meulen \etal~\cite{van_meulen_contactless_2023} also classify sleep stages into four classes, but do so using camera-derived inter-pulse intervals (IPIs). The authors directly apply a deep learning model, previously trained on heart-rate variability (HRV) features derived from the electrocardiogram (ECG), to HRV features computed from a camera-derived IPI time series. This achieves an accuracy of 68\% and a Cohen's $\kappa$ of 0.49 across 46 healthy participants.
\section{Methods}
\subsection{Data acquisition}
In this work, we use video as the input data and polysomnogram recordings as the source of sleep stage labels collected from 50 healthy adult volunteers (18+) with no previous diagnoses of sleep disorders. We report age, weight, sex and Fitzpatrick skin type for the study population in \Cref{table:demographics}. Fitzpatrick skin type~\cite{fitzpatrick_validity_1988} is a measure of skin pigmentation which classifies skin phototypes into one of six categories ranging from I (very pale) to VI (very dark).
\begin{table}[htb]
\caption{Sleep study population demographics.}
\begin{tabular}{@{}ll@{}}
\toprule
Variable              & Value                                \\ \midrule
Age$^1$                   & 37{\small$\pm$15} years\\
Weight$^1$                   & 70.4{\small$\pm$12.9} kg\\
Sex        &                                      \\
\quad Male                  & 22 volunteers                                  \\
\quad Female                & 28 volunteers                                  \\
Fitzpatrick skin type$^{2}$ & I:\@13, II:\@21, III:\@12, IV:\@3, VI:\@1 \\ \bottomrule
\end{tabular}\\
\label{table:demographics}
{\footnotesize $^1$Mean ($\pm$ std). $^2$Skin type: No. volunteers.} \\
\end{table}

\begin{figure}[htbp]
  \centering
   \includegraphics[width=0.85\linewidth]{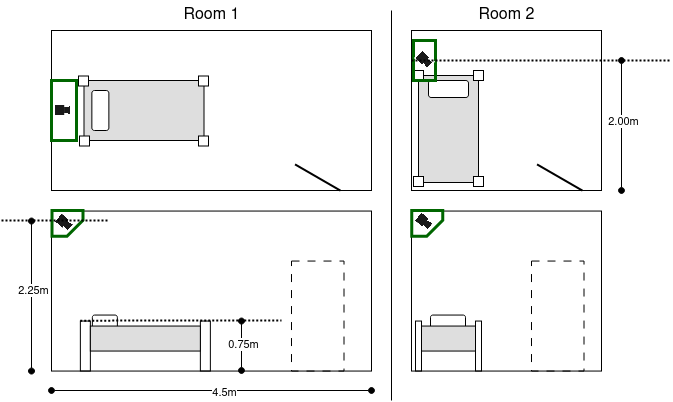}
   \caption{Room and camera layouts used for data acquisition.}\label{fig:layout}
\end{figure}

Each recording took place overnight in one of two experiment rooms. Video data was captured in each room at 20 FPS using a single-channel 850 nm NIR video camera. Both rooms also contained LED illuminators centred at the same wavelength and placed adjacent to the cameras. \Cref{fig:layout} shows the most common bed orientations and the fixed camera positions used in each room.

Video polysomnography data was acquired using a SOMNOscreen$^{\text{TM}}$ Plus device, which recorded a typical vPSG montage which included respiratory inductance plethysmography (RIP), electrooculogram (EOG) and EEG signals and video data, using a separate camera set-up. Each recording was manually scored according to AASM guidelines in 30-second epochs by an experienced sleep physiologist.

\subsection{Optical flow-based activity measurement}
\begin{figure*}[t]
  \centering
   \includegraphics[width=1.0\linewidth]{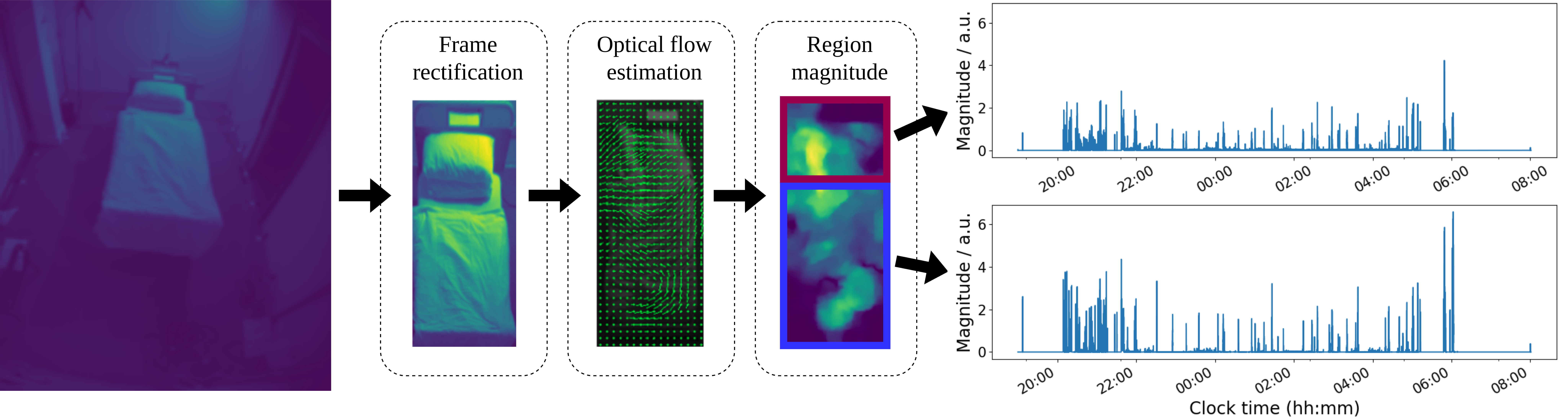}
   \caption{Extracting 1D activity signals from video using optical flow. A homography transformation is applied to each frame followed by cropping to the bed region, to obtain a consistent, virtual viewpoint. The magnitudes of optical flow vectors calculated between frames are averaged over the upper and lower bed regions to produce continuous activity signals for each region.}
   \label{fig:videoactigraphy}
\end{figure*}

From the video camera, we derive continuous measures of subject activity using optical flow~\cite{horn_determining_1981}. This process is illustrated in \Cref{fig:videoactigraphy}. First, we apply an image homography transformation~\cite{hartley_multiple_2003} which provides a consistent, virtual viewpoint of the bed region for different camera positions. This technique has previously been used to improve the robustness of sleep pose detection~\cite{mohammadi_transfer_2021}. We then compute optical flow between transformed frames using the Dense Inverse Search algorithm~\cite{kroeger_fast_2016}. From the 20 FPS video, we estimate the optical flow field between every 5th frame, then average over the previous four estimated fields, giving a field estimation frequency of 4 Hz.

From the optical flow field time series, we derive two simple activity signals by averaging the flow field magnitude in the upper third and lower two thirds of the rectified view, which correspond to approximate head and lower body regions. This gives a simple measure of the spatial scale of movements, since whole-body movements such as sleep posture changes will result in high activity for both signals.

\subsection{Motion features}
\label{section:motion}
From our two video-derived activity signals, we compute a total of 20 time-series features for each epoch, listed in \Cref{table:motionfeatures}. These features broadly fall into two categories: integral features, which quantify the amount of motion that occurred within a time window centred on the epoch, and counter-based features, which quantify the elapsed time since a movement occurred. Our counter-based features are inspired by those of Nochino \etal~\cite{nochino_sleep_2019}.

\begin{table}[htbp]
\centering
\caption{Motion features derived from the 1D optical flow activity signals for each sleep epoch. Features are computed from the upper and lower bed region signals separately.}
\begin{tabular}{ll}
\hline
\textbf{Count}      & \textbf{Feature description                                                                                                     }           \\ \hline
\textbf{14} & \textbf{Integral features}                                                                                                   \\
6          & \begin{tabular}[c]{@{}l@{}}\,30-second Gaussian-weighted sums, time-shifted\\ \,by (-30, 0.0, 30.0) seconds.\end{tabular}      \\
6          & \begin{tabular}[c]{@{}l@{}}\,180-second Gaussian-weighted sums, time-shifted\\ \,by (-180.0, 0.0, 180.0) seconds.\end{tabular} \\
2          & \begin{tabular}[c]{@{}l@{}}\,20-minute Gaussian weighted sums.\end{tabular}\\
\textbf{6} & \textbf{Counter-based features}                                                                                                \\
4          & \begin{tabular}[c]{@{}l@{}}\,Time elapsed since activity signal was above\\ \,a threshold $\in \{0.1, 1.0\}$.\end{tabular}                    \\
2          & \begin{tabular}[c]{@{}l@{}}\,Time elapsed since activity signal was above\\ \,a threshold (1.0) for 3 seconds.\end{tabular} \\ \hline
\end{tabular}\label{table:motionfeatures}
\end{table}
\pagebreak
\subsection{Video-based vital-sign monitoring}\label{section:ncvs}
Many prior works have shown that both the heart rate~\cite{verkruysse_remote_2008, lewandowska_measuring_2011, poh_advancements_2011, wang_algorithmic_2017} and breathing rate~\cite{nakajima_development_2001, li_non-contact_2014, wang_algorithmic_2022} can be measured using video cameras. These methods have been validated in a number of clinical settings e.g.~\cite{trumpp_camera-based_2018, villarroel_non-contact_2020, jorge_non-contact_2022}. The recent study of van Gastel \etal~\cite{van_gastel_camera-based_2021} has also demonstrated the feasibility of continuous vital-sign monitoring during sleep.

In this work, we use components of Oxevision Vital Signs medical device software \cite{hutchinson_method_2023} to measure the heart rate and breathing rate from video data. For both vital signs, this produces 1 Hz estimates and binary Signal Quality Indices (SQIs, \cite{li_robust_2007}), which indicate the software's assessment of the estimate's quality.

The software estimates both vital signs through a process of region of interest (ROI) selection, spatio-temporal filtering, cardio-respiratory signal extraction and data fusion. First, ROIs are identified by selecting areas in the image containing high variance that is periodic in nature~\cite{hutchinson_method_2023}. These ROIs are then spatially smoothed with a median filter and band-pass filtered using pass-bands of 40--130 BPM and 8--39 BRPM for the heart and breathing rates respectively. The cardio-respiratory signals are extracted from the ROIs using principal component analysis as done in Villaroel \etal~\cite{villarroel_non-contact_2020}, with the heart and breathing rates identified from the dominant frequency components. SQIs are derived for each vital sign from heuristics including inter-beat interval consistency. These SQIs are typically low when the volunteer is active, e.g. out of bed, or absent from the room.

\begin{figure}[thp]
  \centering
   \includegraphics[width=0.75\linewidth]{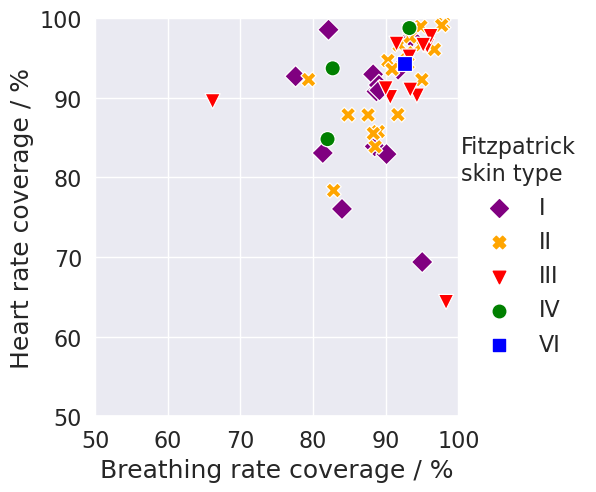}
   \caption{Coverage of video-derived vital-sign estimates.}\label{fig:vitalscoverage}
\end{figure}

We define the coverage of each vital sign as the percentage of time for which the SQI is high when the volunteer is in bed. Using this definition, \Cref{fig:vitalscoverage} shows the coverage for each vital sign across recordings in the dataset. Coverage statistics computed across participants are given in \Cref{table:vitalscoverage}.

\begin{figure}[htbp]
  \centering
   \includegraphics[width=1.0\linewidth]{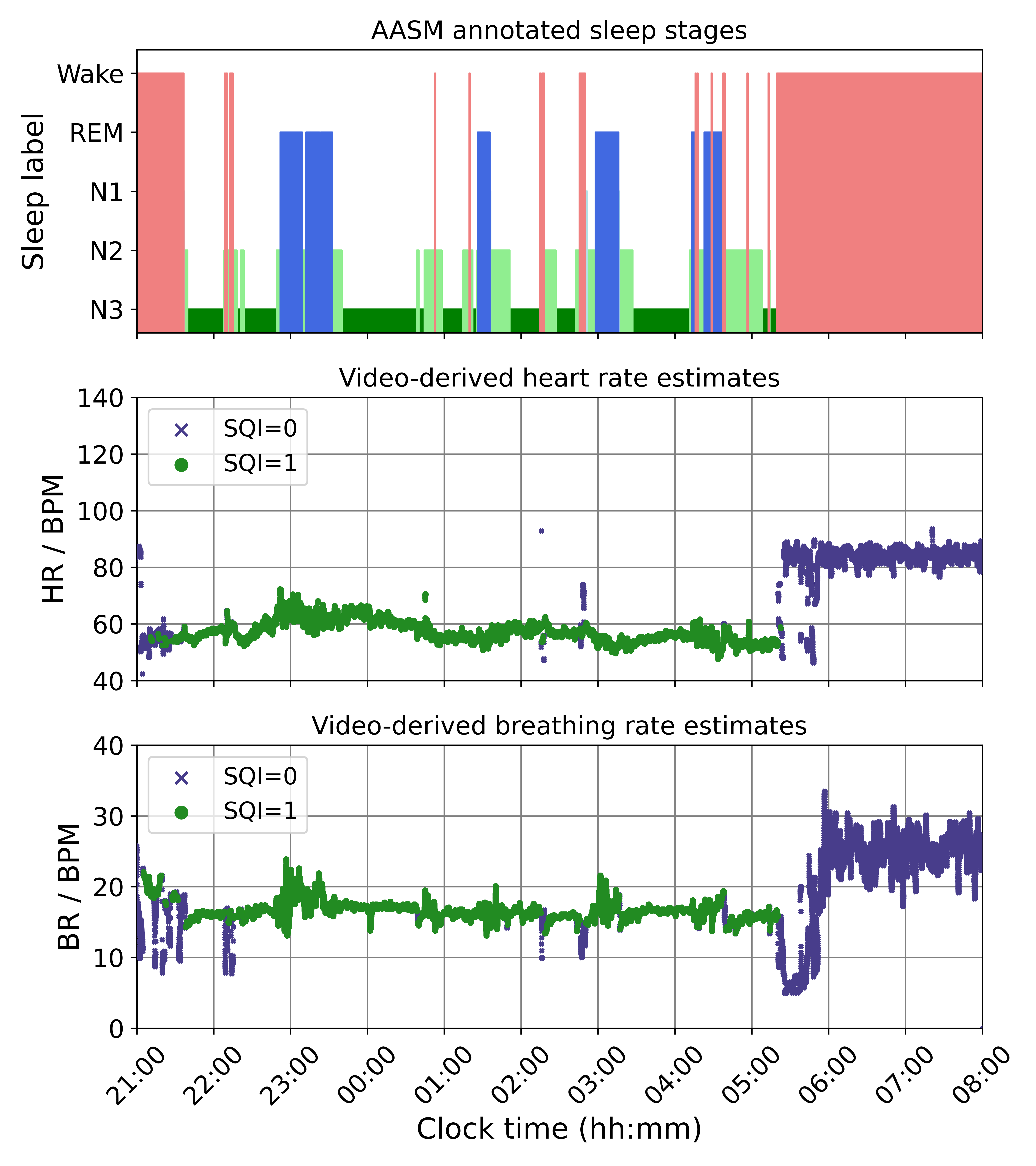}
   \caption{Example AASM sleep stages and video-derived heart rate and breathing estimates for a night's recording from the video dataset.}\label{fig:cameravitals}
\end{figure}

\Cref{fig:cameravitals} shows an example of heart rate and breathing rate time-series estimated from near-infrared video over a night's recording along with AASM-annotated sleep stages. In this example, we can observe increased variance in the breathing rate around REM cycles, indicating how sleep state information can be encoded in the cardio-respiratory time series.

\begin{table}[htbp]
\centering
\caption{Coverage statistics for video-derived vital signs.}
\begin{tabular}{@{}llll@{}}
\toprule
                             & Mean $\pm$ Std  & Min  & Max  \\ \midrule
BR coverage (\%) & 91.0{\small$\pm$7.5} & 66.2 & 98.3 \\
HR coverage (\%)     & 90.0{\small$\pm$6.0} & 64.3 & 99.3 \\ \bottomrule
\end{tabular}\label{table:vitalscoverage}
\end{table}

\subsection{Vital-sign representation learning}\label{section:vitalsigns}
State-of-the-art EEG-based sleep staging performance has been achieved using supervised deep learning methods trained with large datasets, often containing thousands of PSG recordings~\cite{phan_automatic_2022}. These datasets e.g. SHHS~\cite{quan_sleep_1997}, required multiple years to collect across several data collection centres.

By training a model on a large, source dataset and fine-tuning on a smaller target dataset it is often possible to achieve greater performance than training directly on the target dataset. This approach has successfully been applied in the context of wearable sleep staging~\cite{radha_deep_2021}. The authors first train a model to classify sleep on a large ECG dataset before fine-tuning it on a much smaller target photoplethysmography (PPG) dataset. This is shown to give improved performance over models trained solely on the target PPG dataset.

We use the SHHS dataset~\cite{quan_sleep_1997} to learn representations from heart rate and breathing rate time-series, acquired using wearable sensors, which can then be applied to similar time series obtained from video data. The SHHS dataset is substantially larger than our target video dataset, with polysomnogram recordings and accompanying AASM sleep stage labels from around 6000 participants.

\begin{figure}[t]
  \centering
   \includegraphics[width=1.0\linewidth]{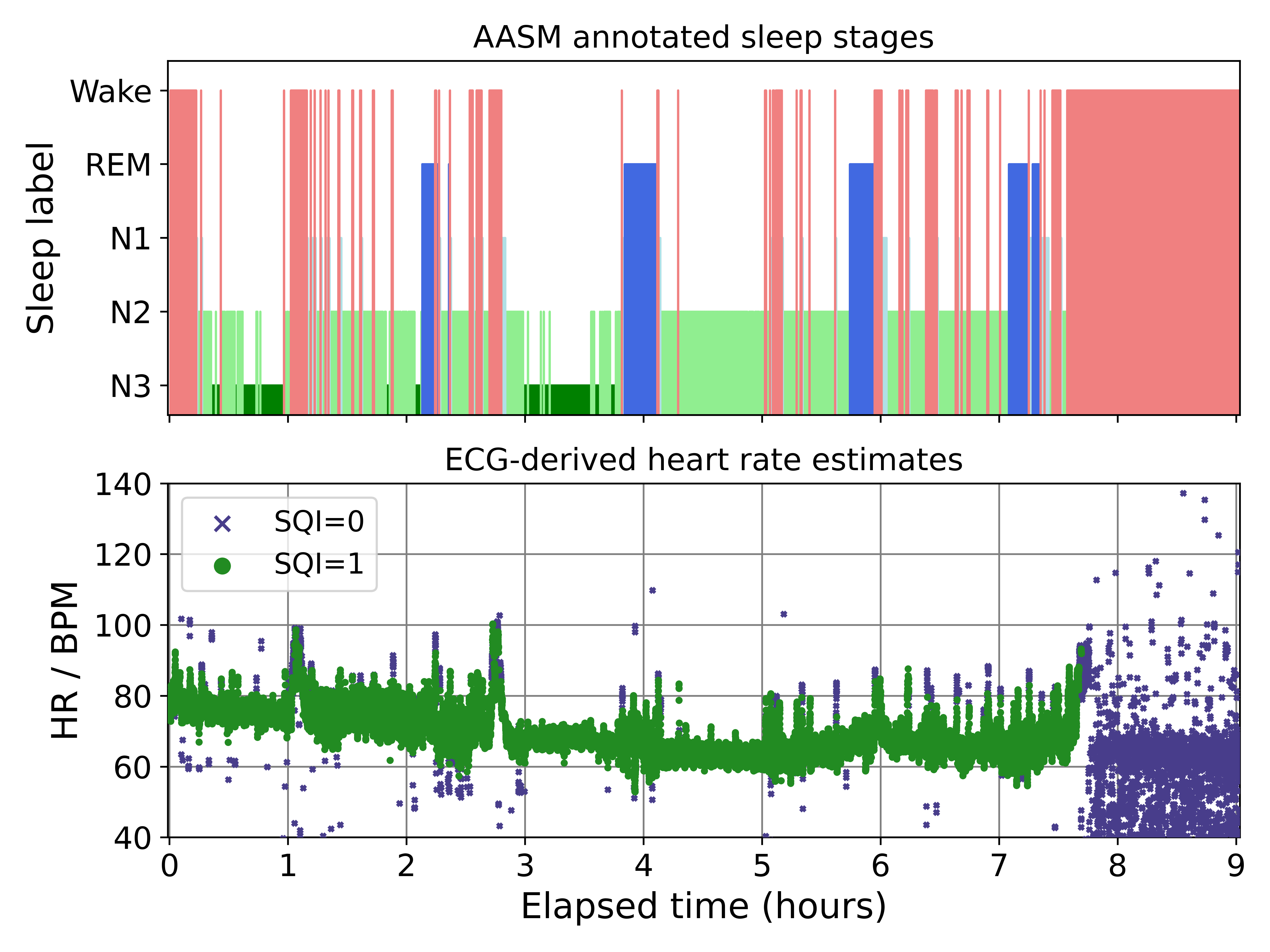}
   \caption{Example AASM sleep stages and ECG-derived heart rate estimates and SQI values from an SHHS polysomnogram recording.}\label{fig:shhsecg}
\end{figure}

\textbf{Training inputs.}
Firstly, heart rate (HR) and breathing rate (BR) time-series for the SHHS dataset were estimated from the ECG and RIP signals respectively. An SQI was also calculated for each estimate. More information on our estimation process is given in \cref{section:shhsestimation}. \Cref{fig:shhsecg} shows synchronised AASM sleep stage labels and ECG-derived heart rate estimates for an example recording from the SHHS dataset. We see that there is a long period of low signal quality (SQI=0) at the end of the recording after the participant has woken up. Periods of low signal quality are strongly correlated with wakefulness, since they are often induced by motion artefacts.

\textbf{SQI Filtering.} To improve performance, we set low quality sections of the time series to zero before passing them to the model. By zeroing out these sections, we still allow the model to know which periods are of low quality whilst preventing the model from overfitting to signal noise.

\begin{figure}[htb]
  \centering
   \includegraphics[width=1.0\linewidth]{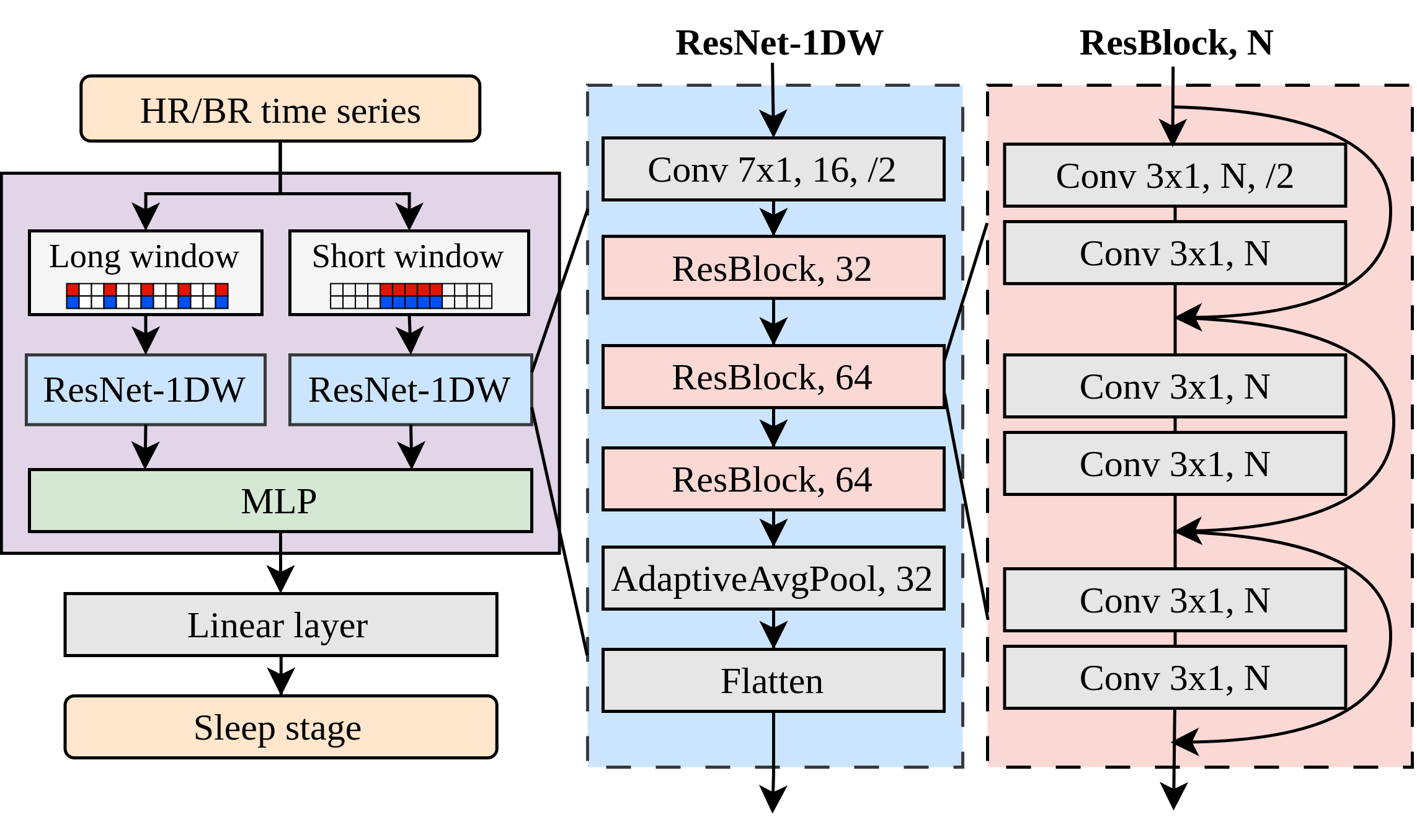}
   \caption{Sleep staging model architecture used to learn vital-sign representations from the SHHS dataset. When applying the model to video data, we use the outputs of the MLP (purple box) as input features to an alternative classifier which also incorporates motion information.}\label{fig:modelarchitecture}
\end{figure}

\textbf{Model architecture.} \Cref{fig:modelarchitecture} illustrates our neural network architecture. The model uses two input streams to classify each epoch: a high-frequency 5-minute HR/BR window sampled at 1 Hz and a lower frequency 50-minute window sampled at 0.1 Hz, both centred around the target epoch for classification. Each window of data is passed to a 1D ResNet \cite{he_deep_2016} model, which transforms the two-channel input time series into a feature vector. Our ResNet architecture broadly follows the design of the original paper, except that 2D convolutions are replaced with their 1D equivalents. The feature vectors are concatenated and passed to a multi-layer perceptron (MLP), which mixes short-term and long-term information from the windows to produce a single feature vector. Finally, a linear classification layer is applied to the feature vector output of the MLP to obtain sleep stage probabilities.

\textbf{Model training.}
The network was trained end-to-end using back-propagation, minimising the cross-entropy loss function against the annotated AASM (5-class) sleep stages. Further model implementation and training details are given in \Cref{section:implementation}. After training, we expect sleep stages to be linearly separable from the output features of the MLP, thus making them effective inputs to any alternative classification model. We use the trained model as a feature extractor by taking the outputs from the MLP layer.

\subsection{Transfer learning}\label{section:transferlearning}
From our video-derived heart rate and breathing rate time series, we use the trained feature extractor to produce `deep' vital-sign features. These are used in conjunction with the motion features described in \Cref{section:motion} as inputs to a random forest classifier which is trained to classify sleep stages on an epoch-by-epoch basis. A random forest classifier was chosen for its combination of low computational cost and robustness to overfitting.
\subsection{Evaluation}
The model was trained and evaluated using k-fold cross-validation, using 10 non-overlapping folds each containing 5 recordings. Performance on each fold was evaluated using a model trained on the remaining 9 folds. Results were then aggregated over all folds.
\section{Results and Discussion}
In this section, we report performance in terms of multi-class classification accuracy and Cohen's kappa between model output classifications and annotated sleep stages. A comparison with existing video-based four-class sleep staging algorithms is given in \Cref{table:comparison}.
\begin{table}[htbp]
\caption{Performance comparison between our deep transfer learning method and prior video-based four-class sleep staging work.}
\centering
\begin{tabular}{@{}llll@{}}
\toprule
Method     & N$^1$ & Cohen's $\kappa$ & Accuracy (\%) \\ \midrule
Nochino \etal \cite{nochino_sleep_2019} & 6 & 0.19{\small$\pm$0.04} & 40.5{\small$\pm$2.2} \\
van Meulen \etal\cite{van_meulen_contactless_2023} & 46 & 0.49{\small$\pm$0.13} & 67.9{\small$\pm$8.7} \\
Ours & 50 & \textbf{0.61}{\small$\pm$0.15} & \textbf{73.4}{\small$\pm$9.6} \\ \bottomrule
\end{tabular}\label{table:comparison}
\raggedright
{\footnotesize $^1$Study population size. Mean ($\pm$ std) across recordings.}
\end{table}

\begin{figure}[htb]
  \centering
   \includegraphics[width=1.0\linewidth]{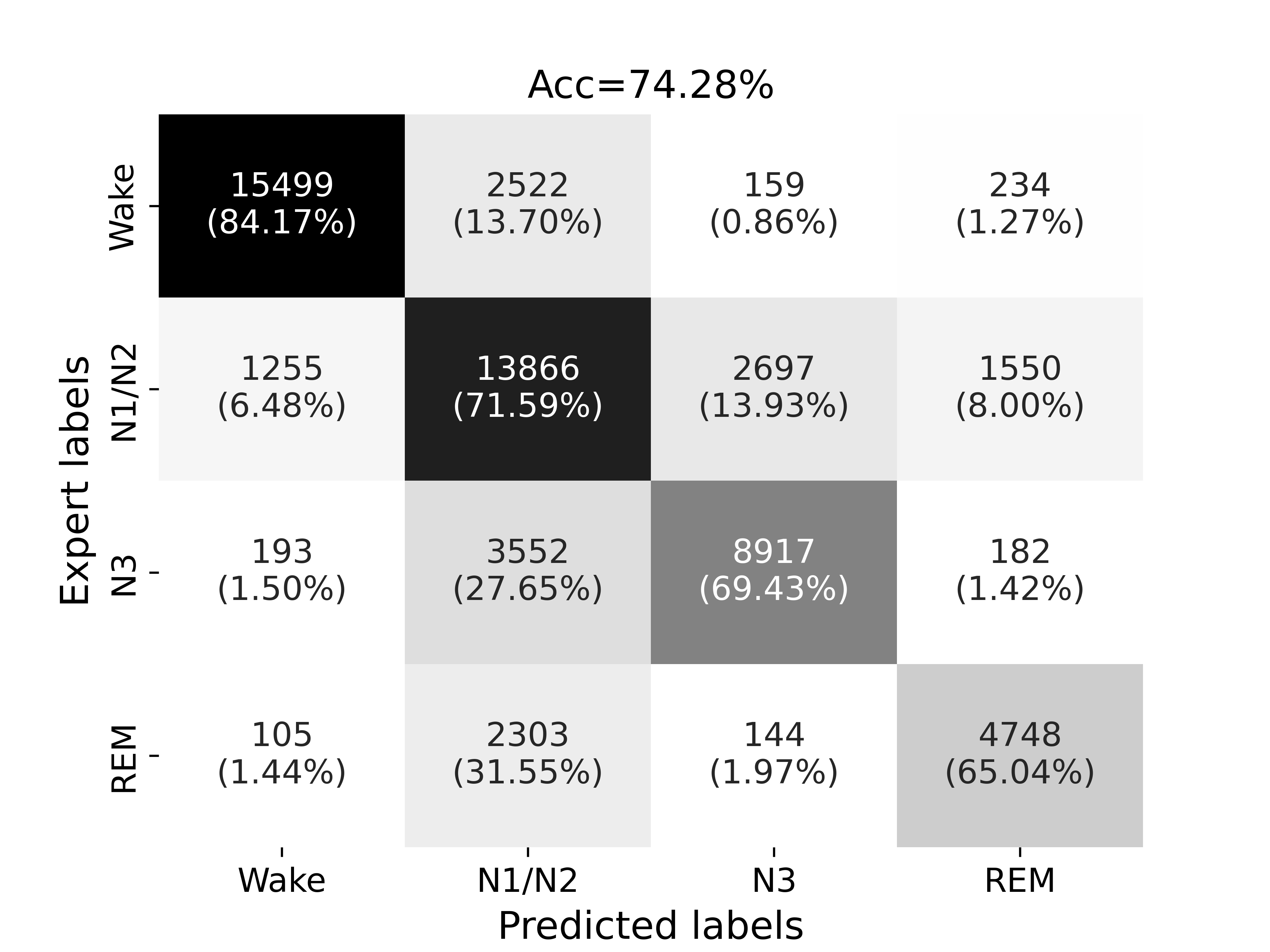}
   \caption{Four-class sleep staging confusion matrix across the dataset using our best performing method. Row-wise percentages are given in brackets, with values on the diagonal indicating class-wise sensitivities.}\label{fig:deep_conf_mat}
\end{figure}

\Cref{fig:deep_conf_mat} shows the total test confusion matrix between scorer and model labels summed over all recordings.

\begin{figure}[htb]
  \centering
   \includegraphics[width=0.9\linewidth]{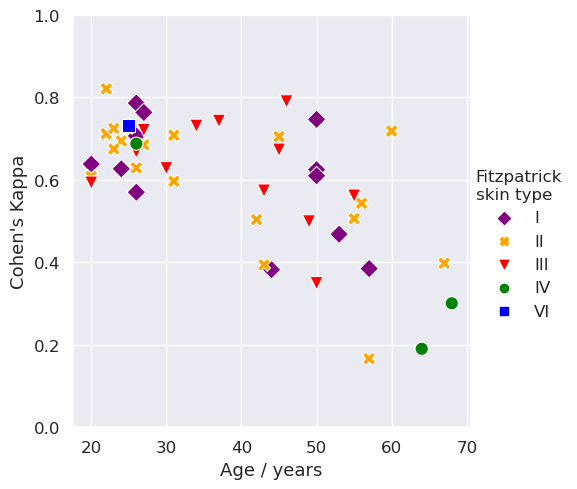}
   \caption{Scatter plot of Cohen's kappa values against age and Fitzpatrick skin type.}\label{fig:agescatter}
\end{figure}

\textbf{Performance with age.}
\Cref{fig:agescatter} shows the variation in Cohen's kappa values with age. Here we observe a clear negative correlation between age and the Cohen's kappa statistic, findings which mirror those from prior ECG-based sleep staging work~\cite{radha_sleep_2019}. In older adults, common measures of autonomic function such as HRV are known to decrease with age \cite{ziegler_assessment_1992}, making sleep stages more difficult to distinguish when using cardio-respiratory inputs.

\begin{figure}[htbp]
  \centering
   \includegraphics[width=1.0\linewidth]{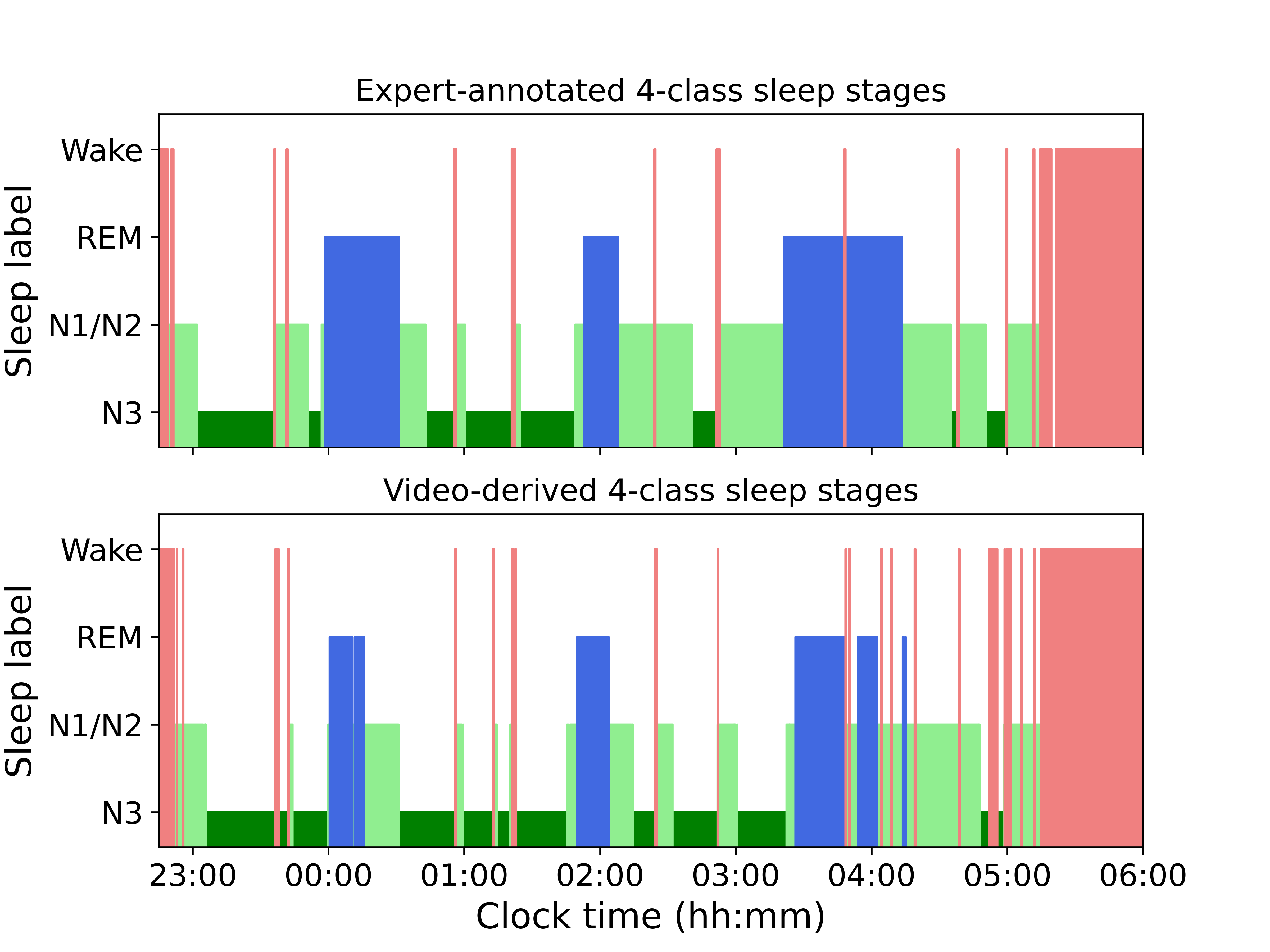}
   \caption{Model and scorer four-class hypnograms corresponding to the median Cohen's kappa statistic (0.63).}\label{fig:medianhypnogram}
\end{figure}

\textbf{Example hypnogram output.} \Cref{fig:medianhypnogram} shows an example four-class sleep hypnogram produced by our model. This example corresponds to the median Cohen's kappa obtained across recordings. There is good visual agreement in the sleep architecture: the model correctly identifies all three REM cycles, sleep onsets and offsets, and most brief arousals during the night. Hypnograms corresponding to the minimum and maximum Cohen's kappa are given in \Cref{section:extrahypnograms}.

\textbf{Performance with classification strategy.}
\Cref{table:sleepstratification} indicates the performance of our model using alternative sleep stage classification strategies. The random forest classifier was re-trained for each strategy against the target labels which are derived from the AASM labels accordingly.
\begin{table}[htb]
\centering
\caption{Performance comparison across Sleep--Wake (SW), Wake--NREM--REM (3-class) and Wake--N1/N2--N3--REM (4-class) classification strategies.}
\begin{tabular}{@{}lllll@{}}
\toprule
              & Cohen's $\kappa$ & Acc. (\%) & TPR$^{1}$ (\%) & TNR$^{2}$ (\%) \\ \midrule
SW    & 0.72{\small$\pm$0.24}    & 91.6{\small$\pm$6.9} & 94.2{\small$\pm$5.3}& 81.0{\small$\pm$25.0}    \\
3-class & 0.68{\small$\pm$0.18}    & 84.9{\small$\pm$7.1} & -                & -                \\
4-class & 0.61{\small$\pm$0.15}    & 73.4{\small$\pm$9.6} & -                & -                \\ \bottomrule
\end{tabular}
\raggedright
{\footnotesize $^1$True positive rate i.e. sensitivity. $^2$True negative rate i.e. specificity.}\label{table:sleepstratification}
\end{table}

\subsection{Ablation studies}
\textbf{Addition of motion and deep features.}
In \Cref{table:featureablation} we assess the impact of including motion features and of using our deep vital-sign features over a simpler baseline of hand-designed features (see \Cref{section:baseline}). We observe that both the deep features and motion features are important to the overall performance.
\begin{table}[t]
\caption{Ablation results showing the effect of adding motion features (M) and of using deep features (DF) extracted from the vital signs compared with a simple feature engineering (FE) baseline.}
\centering
\begin{tabular}{@{}lll@{}}
\toprule
Method     & Cohen's $\kappa$ & Accuracy (\%) \\ \midrule
FE & 0.38{\small$\pm$0.20} & 57.6{\small$\pm$13.6} \\
FE + M & 0.56{\small$\pm$0.17} & 70.2{\small$\pm$10.2} \\
DF & 0.50{\small$\pm$0.16} & 65.5{\small$\pm$10.5} \\
DF + M & 0.61{\small$\pm$0.15} & 73.4{\small$\pm$9.6} \\ \bottomrule
\end{tabular}\label{table:featureablation}
\end{table}

\textbf{SQI filtering.}
\Cref{table:sqiablation} indicates the performance improvement obtained from our SQI filtering approach. By zeroing out low quality sections of the time series in both training and testing we achieve an appreciable improvement in our results.
\begin{table}[t]
\caption{Ablation results showing the effect of SQI filtering.}
\centering
\begin{tabular}{@{}lll@{}}
\toprule
Method     & Cohen's $\kappa$ & Accuracy (\%) \\ \midrule
Raw time-series & 0.59{\small$\pm$0.16} & 72.4{\small$\pm$10.0} \\
SQI filtering & 0.61{\small$\pm$0.15} & 73.4{\small$\pm$9.6} \\ \bottomrule
\end{tabular}\label{table:sqiablation}
\end{table}
\section{Conclusions and Future Work}
In this work, we have introduced a novel transfer learning approach to video-based sleep stage classification which uses non-contact measurements of the heart rate, breathing rate and activity measured using a near-infrared camera. This approach achieves state-of-the-art performance on video-based sleep staging on a healthy study population.

Performance is likely to be further improved through the use of an explicit sequence--sequence architecture such as a Transformer \cite{vaswani_attention_2017}, which has led to superior performance in PSG-based sleep staging \cite{phan_sleeptransformer_2022}. Using deep motion features, such as the intermediate outputs of a sleep pose detection model~\cite{mohammadi_transfer_2021}, may also improve classification performance over the relatively simple measures of activity used in this work.

In future work, we intend to investigate the performance of our approach across a broader demographic range and in a sleep-disordered population.

\section*{Acknowledgements}
This work was supported by the EPSRC Centre for Doctoral Training in Autonomous Intelligent Machines and Systems [EP/S024050/1] and funded by Oxehealth Ltd.

We gratefully acknowledge the National Sleep Research Resource~\cite{zhang_national_2018} for granting access to the SHHS dataset.

\appendix
\section{SHHS vital-sign estimation}
\label{section:shhsestimation}
\textbf{Heart rate:} We estimated the heart rate from the 128 Hz ECG signal available from each PSG recording. First, we applied the Pan-Tompkins algorithm~\cite{pan_real-time_1985} to produce a QRS detection time series. We then applied an FFT to 9-second rolling windows, taking the frequency of the main spectral peak as our estimate. An SQI was calculated for each heart rate estimate given by:
\begin{equation}
    \text{SQI}_t^{\text{ECG}} = \begin{cases}
        1& \text{if } \text{\footnotesize \begin{math}\lvert {\text{HR}}^{\text{ECG}}_t - {\text{HR}}^{\text{PPG}}_t \rvert\end{math}} \leq 3\text{\,bpm}\\
    0,              & \text{otherwise}
\end{cases}
\end{equation}
i.e. our SQI is 1 when there is good agreement between our ECG-derived heart rate, {\small$\text{HR}^{\text{ECG}}_t$}, and a PPG-derived heart rate, {\small${\text{HR}}^{\text{PPG}}_t$}, which is available from the SHHS dataset.

\textbf{Breathing rate:} We used a peak counting approach to estimate the breathing rate from both the abdomen and thorax RIP waveforms, using 30-second rolling windows. We then used the thorax-derived estimate as our breathing rate and used the disagreement between the two sources to derive a binary SQI:
\begin{equation}
    \text{SQI}_t^{\text{RIP}} = \begin{cases}
        1& \text{if } \text{\footnotesize \begin{math}\lvert {\text{BR}}^{\text{Abd}}_t - {\text{BR}}^{\text{Thor}}_t \rvert\end{math}} \leq 3\text{\,brpm}\\
    0,              & \text{otherwise}
\end{cases}
\end{equation}
where ${\text{BR}}^{\text{Abd}}_t$ and ${\text{BR}}^{\text{Thor}}_t$ are the time-series estimates derived from the abdomen and thorax respectively.

\section{Further model implementation details}
\label{section:implementation}
Each ResNet model produced a flattened 1D feature vector of length 2048. Our MLP therefore had an input dimension of 4096, given by the concatenation of the short- and long-window feature vectors.  The MLP used a single hidden layer with a hidden dimension of length 100. The output of the MLP i.e. the `deep' feature vector had length 32. This was chosen to be small and similar in size to the motion feature vector, to mitigate overfitting and encourage a balanced weighting of the two feature sources in the transfer learning phase.

To improve training stability, we normalised the input time series passed to the model by dividing by the mean and standard deviation of the heart and breathing rates calculated across the entire SHHS dataset. We used the Adam~\cite{kingma_adam_2014} optimiser with default hyper-parameters and employed early stopping after three epochs without an improvement in validation accuracy.

\section{Feature engineering baseline}
\label{section:baseline}
Our baseline consisted of 6 features extracted from the heart rate and breathing rate time series for each epoch: means over 3-minute rolling windows and standard deviations calculated using 3-minute and 10-minute rolling windows. Each time series was normalised by the 90th percentile value and filtered using a difference of Gaussians filter before computing features as done in the wearable sleep staging work of Walch \etal~\cite{walch_sleep_2019}.

\section{Additional model hypnograms}
\label{section:extrahypnograms}
\begin{figure}[htbp]
  \centering
   \includegraphics[width=1.0\linewidth]{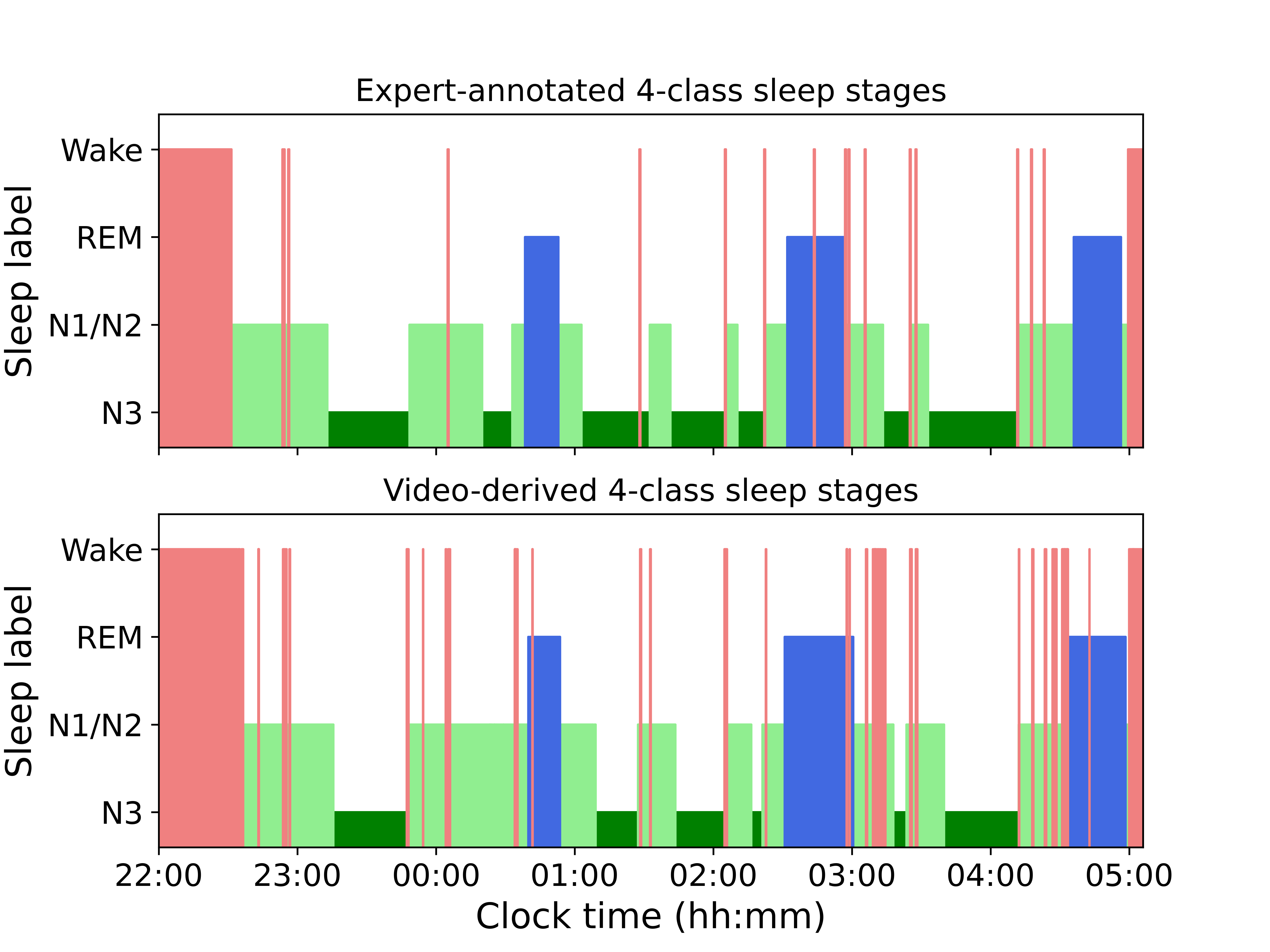}
   \caption{Four-class hypnograms corresponding to the highest Cohen's kappa statistic between model and scorer (0.82).}
   \label{fig:highesthypnogram}
\end{figure}

\Cref{fig:highesthypnogram} shows model and scorer four-class sleep hypnograms for the recording with the highest agreement as measured by Cohen's kappa.

\begin{figure}[hb]
  \centering
   \includegraphics[width=1.0\linewidth]{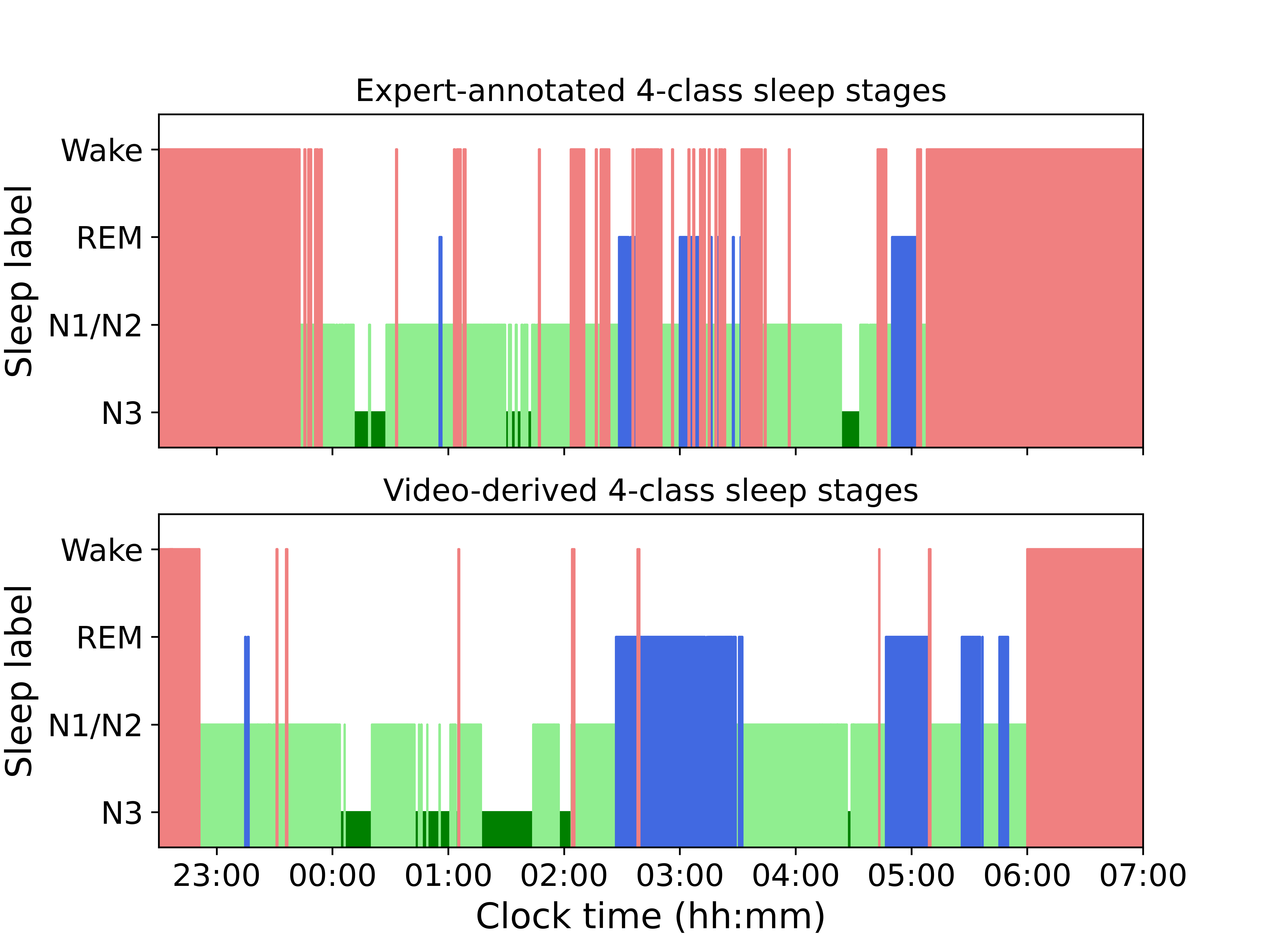}
   \caption{Four-class hypnograms corresponding to the lowest Cohen's kappa statistic between model and scorer (0.17).}
   \label{fig:lowesthypnogram}
\end{figure}

\Cref{fig:lowesthypnogram} shows the hypnograms obtained for the recording with the lowest agreement. In their report, the sleep physiologist noted that there was a `fuzzy signal on [many of the EEG channels] throughout, so sleep staging was difficult and not precise'. Measurement factors such as electrode placement and electrical interference are a common source of scorer uncertainty in conventional sleep staging~\cite{van_gorp_certainty_2022}.

{\small
\bibliographystyle{ieee_fullname}

\begin{thebibliography}{10}\itemsep=-1pt

    \bibitem{agnew_first_1966}
    H.~W. Agnew, W.~B. Webb, and R.~L. Williams.
    \newblock The first night effect: an {EEG} study of sleep.
    \newblock {\em Psychophysiology}, 2(3):263--266, Jan. 1966.
    
    \bibitem{cohen_coefficient_1960}
    Jacob Cohen.
    \newblock A {Coefficient} of {Agreement} for {Nominal} {Scales}.
    \newblock {\em Educational and Psychological Measurement}, 20(1):37--46, Apr. 1960.
    
    \bibitem{cuppens_automatic_2010}
    Kris Cuppens, Lieven Lagae, Berten Ceulemans, Sabine Van~Huffel, and Bart Vanrumste.
    \newblock Automatic video detection of body movement during sleep based on optical flow in pediatric patients with epilepsy.
    \newblock {\em Medical \& Biological Engineering \& Computing}, 48(9):923--931, Sept. 2010.
    
    \bibitem{danker-hopfe_interrater_2009}
    Heidi Danker-Hopfe, Peter Anderer, Josef Zeitlhofer, Marion Boeck, Hans Dorn, Georg Gruber, Esther Heller, Erna Loretz, Doris Moser, Silvia Parapatics, Bernd Saletu, Andrea Schmidt, and Georg Dorffner.
    \newblock Interrater reliability for sleep scoring according to the {Rechtschaffen} \& {Kales} and the new {AASM} standard.
    \newblock {\em Journal of Sleep Research}, 18(1):74--84, Mar. 2009.
    
    \bibitem{delrosso_video-polysomnographic_2019}
    Lourdes~M DelRosso, Caroline~V Jackson, Kimberly Trotter, Oliviero Bruni, and Raffaele Ferri.
    \newblock Video-polysomnographic characterization of sleep movements in children with restless sleep disorder.
    \newblock {\em Sleep}, 42(4), Apr. 2019.
    
    \bibitem{fitzpatrick_validity_1988}
    T.~B. Fitzpatrick.
    \newblock The validity and practicality of sun-reactive skin types {I} through {VI}.
    \newblock {\em Archives of Dermatology}, 124(6):869--871, June 1988.
    
    \bibitem{hartley_multiple_2003}
    Richard Hartley and Andrew Zisserman.
    \newblock {\em Multiple view geometry in computer vision}.
    \newblock Cambridge university press, 2003.
    
    \bibitem{he_deep_2016}
    Kaiming He, Xiangyu Zhang, Shaoqing Ren, and Jian Sun.
    \newblock Deep {Residual} {Learning} for {Image} {Recognition}.
    \newblock In {\em Proceedings of the {IEEE} {Conference} on {Computer} {Vision} and {Pattern} {Recognition}}, pages 770--778, 2016.
    
    \bibitem{heinrich_body_2013}
    Adrienne Heinrich, Xavier Aubert, and Gerard de Haan.
    \newblock Body movement analysis during sleep based on video motion estimation.
    \newblock In {\em 2013 {IEEE} 15th {International} {Conference} on e-{Health} {Networking}, {Applications} and {Services} ({Healthcom} 2013)}, pages 539--543, Oct. 2013.
    
    \bibitem{heinrich_robust_2014}
    Adrienne Heinrich, Di Geng, Dmitry Znamenskiy, Jelte~Peter Vink, and Gerard de Haan.
    \newblock Robust and {Sensitive} {Video} {Motion} {Detection} for {Sleep} {Analysis}.
    \newblock {\em IEEE Journal of Biomedical and Health Informatics}, 18(3):790--798, May 2014.
    \newblock Conference Name: IEEE Journal of Biomedical and Health Informatics.
    
    \bibitem{horn_determining_1981}
    Berthold K.~P. Horn and Brian~G. Schunck.
    \newblock Determining optical flow.
    \newblock {\em Artificial Intelligence}, 17(1):185--203, Aug. 1981.
    
    \bibitem{hudgel_mechanics_1984}
    D.~W. Hudgel, R.~J. Martin, B. Johnson, and P. Hill.
    \newblock Mechanics of the respiratory system and breathing pattern during sleep in normal humans.
    \newblock {\em Journal of Applied Physiology: Respiratory, Environmental and Exercise Physiology}, 56(1):133--137, Jan. 1984.
    
    \bibitem{hutchinson_method_2023}
    Nicholas~Dunkley Hutchinson and Simon Mark~Chave Jones.
    \newblock Method and apparatus for monitoring of a human or animal subject field.
    \newblock US11563920B2, Jan. 2023.
    
    \bibitem{jorge_non-contact_2022}
    João Jorge, Mauricio Villarroel, Hamish Tomlinson, Oliver Gibson, Julie~L. Darbyshire, Jody Ede, Mirae Harford, John~Duncan Young, Lionel Tarassenko, and Peter Watkinson.
    \newblock Non-contact physiological monitoring of post-operative patients in the intensive care unit.
    \newblock {\em npj Digital Medicine}, 5(1):4, Dec. 2022.
    
    \bibitem{kingma_adam_2014}
    Diederik~P. Kingma and Jimmy Ba.
    \newblock Adam: {A} method for stochastic optimization.
    \newblock {\em arXiv preprint arXiv:1412.6980}, 2014.
    
    \bibitem{kroeger_fast_2016}
    Till Kroeger, Radu Timofte, Dengxin Dai, and Luc Van~Gool.
    \newblock Fast {Optical} {Flow} using {Dense} {Inverse} {Search}, Mar. 2016.
    \newblock arXiv:1603.03590 [cs].
    
    \bibitem{kruger_sleep_2014}
    Bjorn Kruger, Anna Vogele, Marouane Lassiri, Lukas Herwartz, Thomas Terkatz, and Andreas Weber.
    \newblock Sleep detection using de-identified depth data.
    \newblock {\em Journal of Mobile Multimedia}, pages 327--342, Dec. 2014.
    
    \bibitem{lewandowska_measuring_2011}
    Magdalena Lewandowska, Jacek Rumiński, Tomasz Kocejko, and Jędrzej Nowak.
    \newblock Measuring pulse rate with a webcam — {A} non-contact method for evaluating cardiac activity.
    \newblock In {\em 2011 {Federated} {Conference} on {Computer} {Science} and {Information} {Systems} ({FedCSIS})}, pages 405--410, Sept. 2011.
    
    \bibitem{li_non-contact_2014}
    Michael~H. Li, Azadeh Yadollahi, and Babak Taati.
    \newblock A non-contact vision-based system for respiratory rate estimation.
    \newblock In {\em 2014 36th {Annual} {International} {Conference} of the {IEEE} {Engineering} in {Medicine} and {Biology} {Society}}, pages 2119--2122, Aug. 2014.
    \newblock ISSN: 1558-4615.
    
    \bibitem{li_robust_2007}
    Q. Li, R.~G. Mark, and G.~D. Clifford.
    \newblock Robust heart rate estimation from multiple asynchronous noisy sources using signal quality indices and a {Kalman} filter.
    \newblock {\em Physiological Measurement}, 29(1):15--32, Dec. 2007.
    \newblock Publisher: IOP Publishing.
    
    \bibitem{long_video-based_2019}
    Xi Long, Renée Otte, Eric van~der Sanden, Jan Werth, and Tao Tan.
    \newblock Video-{Based} {Actigraphy} for {Monitoring} {Wake} and {Sleep} in {Healthy} {Infants}: {A} {Laboratory} {Study}.
    \newblock {\em Sensors}, 19(5):1075, Jan. 2019.
    
    \bibitem{mohammadi_transfer_2021}
    Sara~Mahvash Mohammadi, Shirin Enshaeifar, Adrian Hilton, Derk-Jan Dijk, and Kevin Wells.
    \newblock Transfer {Learning} for {Clinical} {Sleep} {Pose} {Detection} {Using} a {Single} {2D} {IR} {Camera}.
    \newblock {\em IEEE Transactions on Neural Systems and Rehabilitation Engineering}, 29:290--299, 2021.
    
    \bibitem{nakajima_development_2001}
    Kazuki Nakajima, Yoshiaki Matsumoto, and Toshiyo Tamura.
    \newblock Development of real-time image sequence analysis for evaluating posture change and respiratory rate of a subject in bed.
    \newblock {\em Physiological Measurement}, 22(3):N21--N28, Aug. 2001.
    
    \bibitem{nochino_sleep_2019}
    Teruaki Nochino, Yuko Ohno, Takafumi Kato, Masako Taniike, and Shima Okada.
    \newblock Sleep stage estimation method using a camera for home use.
    \newblock {\em Biomedical Engineering Letters}, 9(2):257--265, Apr. 2019.
    
    \bibitem{pan_real-time_1985}
    Jiapu Pan and Willis~J. Tompkins.
    \newblock A {Real}-{Time} {QRS} {Detection} {Algorithm}.
    \newblock {\em IEEE Transactions on Biomedical Engineering}, BME-32(3):230--236, Mar. 1985.
    
    \bibitem{phan_automatic_2022}
    Huy Phan and Kaare Mikkelsen.
    \newblock Automatic sleep staging of {EEG} signals: recent development, challenges, and future directions.
    \newblock {\em Physiological Measurement}, 2022.
    \newblock Publisher: IOP Publishing.
    
    \bibitem{phan_sleeptransformer_2022}
    Huy Phan, Kaare~B Mikkelsen, Oliver Chen, Philipp Koch, Alfred Mertins, and Maarten De~Vos.
    \newblock {SleepTransformer}: {Automatic} {Sleep} {Staging} with {Interpretability} and {Uncertainty} {Quantification}.
    \newblock {\em IEEE Transactions on Biomedical Engineering}, pages 1--1, 2022.
    
    \bibitem{pini_automated_2022}
    Nicolò Pini, Ju~Lynn Ong, Gizem Yilmaz, Nicholas I. Y.~N. Chee, Zhao Siting, Animesh Awasthi, Siddharth Biju, Kishan Kishan, Amiya Patanaik, William~P. Fifer, and Maristella Lucchini.
    \newblock An automated heart rate-based algorithm for sleep stage classification: {Validation} using conventional polysomnography and an innovative wearable electrocardiogram device.
    \newblock {\em Frontiers in Neuroscience}, 16:974192, Oct. 2022.
    
    \bibitem{poh_advancements_2011}
    Ming-Zher Poh, Daniel~J. McDuff, and Rosalind~W. Picard.
    \newblock Advancements in {Noncontact}, {Multiparameter} {Physiological} {Measurements} {Using} a {Webcam}.
    \newblock {\em IEEE Transactions on Biomedical Engineering}, 58(1):7--11, Jan. 2011.
    \newblock Conference Name: IEEE Transactions on Biomedical Engineering.
    
    \bibitem{quan_sleep_1997}
    S.~F. Quan, B.~V. Howard, C. Iber, J.~P. Kiley, F.~J. Nieto, G.~T. O'Connor, D.~M. Rapoport, S. Redline, J. Robbins, J.~M. Samet, and P.~W. Wahl.
    \newblock The {Sleep} {Heart} {Health} {Study}: design, rationale, and methods.
    \newblock {\em Sleep}, 20(12):1077--1085, Dec. 1997.
    
    \bibitem{radha_sleep_2019}
    Mustafa Radha, Pedro Fonseca, Arnaud Moreau, Marco Ross, Andreas Cerny, Peter Anderer, Xi Long, and Ronald~M. Aarts.
    \newblock Sleep stage classification from heart-rate variability using long short-term memory neural networks.
    \newblock {\em Scientific Reports}, 9(1):14149, Oct. 2019.
    
    \bibitem{radha_deep_2021}
    Mustafa Radha, Pedro Fonseca, Arnaud Moreau, Marco Ross, Andreas Cerny, Peter Anderer, Xi Long, and Ronald~M. Aarts.
    \newblock A deep transfer learning approach for wearable sleep stage classification with photoplethysmography.
    \newblock {\em npj Digital Medicine}, 4(1):1--11, Sept. 2021.
    
    \bibitem{schwichtenberg_pediatric_2018}
    A.~J. Schwichtenberg, Jeehyun Choe, Ashleigh Kellerman, Emily~A. Abel, and Edward~J. Delp.
    \newblock Pediatric {Videosomnography}: {Can} {Signal}/{Video} {Processing} {Distinguish} {Sleep} and {Wake} {States}?
    \newblock {\em Frontiers in Pediatrics}, 6:158, June 2018.
    
    \bibitem{shinar_automatic_2001}
    Z. Shinar, A. Baharav, Y. Dagan, and S. Akselrod.
    \newblock Automatic detection of slow-wave-sleep using heart rate variability.
    \newblock In {\em Computers in {Cardiology} 2001. {Vol}.28 ({Cat}. {No}.{01CH37287})}, pages 593--596, Sept. 2001.
    \newblock ISSN: 0276-6547.
    
    \bibitem{stefani_prospective_2015}
    Ambra Stefani, David Gabelia, Thomas Mitterling, Werner Poewe, Birgit Högl, and Birgit Frauscher.
    \newblock A {Prospective} {Video}-{Polysomnographic} {Analysis} of {Movements} during {Physiological} {Sleep} in 100 {Healthy} {Sleepers}.
    \newblock {\em Sleep}, 38(9):1479--1487, Sept. 2015.
    
    \bibitem{stefani_sleep_2020}
    Ambra Stefani and Birgit Högl.
    \newblock Sleep in {Parkinson}’s disease.
    \newblock {\em Neuropsychopharmacology}, 45(1):121--128, Jan. 2020.
    
    \bibitem{trumpp_camera-based_2018}
    Alexander Trumpp, Johannes Lohr, Daniel Wedekind, Martin Schmidt, Matthias Burghardt, Axel~R. Heller, Hagen Malberg, and Sebastian Zaunseder.
    \newblock Camera-based photoplethysmography in an intraoperative setting.
    \newblock {\em BioMedical Engineering OnLine}, 17(1):33, Dec. 2018.
    
    \bibitem{van_gastel_camera-based_2021}
    Mark van Gastel, Sander Stuijk, Sebastiaan Overeem, Johannes~P. van Dijk, Merel~M. van Gilst, and Gerard de Haan.
    \newblock Camera-{Based} {Vital} {Signs} {Monitoring} {During} {Sleep} – {A} {Proof} of {Concept} {Study}.
    \newblock {\em IEEE Journal of Biomedical and Health Informatics}, 25(5):1409--1418, May 2021.
    
    \bibitem{van_gorp_certainty_2022}
    Hans van Gorp, Iris A~M Huijben, Pedro Fonseca, Ruud J~G van Sloun, Sebastiaan Overeem, and Merel~M van Gilst.
    \newblock Certainty about uncertainty in sleep staging: a theoretical framework.
    \newblock {\em Sleep}, 45(8):zsac134, Aug. 2022.
    
    \bibitem{van_meulen_contactless_2023}
    Fokke~B. van Meulen, Angela Grassi, Leonie van~den Heuvel, Sebastiaan Overeem, Merel~M. van Gilst, Johannes~P. van Dijk, Henning Maass, Mark J.~H. van Gastel, and Pedro Fonseca.
    \newblock Contactless {Camera}-{Based} {Sleep} {Staging}: {The} {HealthBed} {Study}.
    \newblock {\em Bioengineering}, 10(1):109, Jan. 2023.
    
    \bibitem{vaswani_attention_2017}
    Ashish Vaswani, Noam Shazeer, Niki Parmar, Jakob Uszkoreit, Llion Jones, Aidan~N. Gomez, Lukasz Kaiser, and Illia Polosukhin.
    \newblock Attention {Is} {All} {You} {Need}.
    \newblock {\em arXiv:1706.03762 [cs]}, Dec. 2017.
    \newblock arXiv: 1706.03762.
    
    \bibitem{veauthier_contactless_2019}
    Christian Veauthier, Juliane Ryczewski, Sebastian Mansow-Model, Karen Otte, Bastian Kayser, Martin Glos, Christoph Schöbel, Friedemann Paul, Alexander~U. Brandt, and Thomas Penzel.
    \newblock Contactless recording of sleep apnea and periodic leg movements by nocturnal 3-{D}-video and subsequent visual perceptive computing.
    \newblock {\em Scientific Reports}, 9(1):16812, Nov. 2019.
    
    \bibitem{verkruysse_remote_2008}
    Wim Verkruysse, Lars~O. Svaasand, and J.~Stuart Nelson.
    \newblock Remote plethysmographic imaging using ambient light.
    \newblock {\em Optics Express}, 16(26):21434--21445, Dec. 2008.
    
    \bibitem{villarroel_non-contact_2020}
    Mauricio Villarroel, João Jorge, David Meredith, Sheera Sutherland, Chris Pugh, and Lionel Tarassenko.
    \newblock Non-contact vital-sign monitoring of patients undergoing haemodialysis treatment.
    \newblock {\em Scientific Reports}, 10(1):18529, Oct. 2020.
    \newblock Number: 1 Publisher: Nature Publishing Group.
    
    \bibitem{walch_sleep_2019}
    Olivia Walch, Yitong Huang, Daniel Forger, and Cathy Goldstein.
    \newblock Sleep stage prediction with raw acceleration and photoplethysmography heart rate data derived from a consumer wearable device.
    \newblock {\em Sleep}, 42(12):zsz180, Dec. 2019.
    
    \bibitem{wang_algorithmic_2022}
    Wenjin Wang and Albertus~C den Brinker.
    \newblock Algorithmic insights of camera-based respiratory motion extraction.
    \newblock {\em Physiological Measurement}, 43(7):075004, July 2022.
    
    \bibitem{wang_algorithmic_2017}
    Wenjin Wang, Albertus~C. den Brinker, Sander Stuijk, and Gerard de Haan.
    \newblock Algorithmic {Principles} of {Remote} {PPG}.
    \newblock {\em IEEE Transactions on Biomedical Engineering}, 64(7):1479--1491, July 2017.
    
    \bibitem{wulterkens_it_2021}
    Bernice~M Wulterkens, Pedro Fonseca, Lieke W~A Hermans, Marco Ross, Andreas Cerny, Peter Anderer, Xi Long, Johannes~P van Dijk, Nele Vandenbussche, Sigrid Pillen, Merel~M van Gilst, and Sebastiaan Overeem.
    \newblock It is {All} in the {Wrist}: {Wearable} {Sleep} {Staging} in a {Clinical} {Population} versus {Reference} {Polysomnography}.
    \newblock {\em Nature and Science of Sleep}, 13:885--897, June 2021.
    
    \bibitem{zhang_national_2018}
    Guo-Qiang Zhang, Licong Cui, Remo Mueller, Shiqiang Tao, Matthew Kim, Michael Rueschman, Sara Mariani, Daniel Mobley, and Susan Redline.
    \newblock The {National} {Sleep} {Research} {Resource}: towards a sleep data commons.
    \newblock {\em Journal of the American Medical Informatics Association: JAMIA}, 25(10):1351--1358, Oct. 2018.
    
    \bibitem{ziegler_assessment_1992}
    Dan Ziegler, G. Laux, K. Dannehl, M. Spüler, H. Mühlen, P. Mayer, and F.a. Gries.
    \newblock Assessment of {Cardiovascular} {Autonomic} {Function}: {Age}-related {Normal} {Ranges} and {Reproducibility} of {Spectral} {Analysis}, {Vector} {Analysis}, and {Standard} {Tests} of {Heart} {Rate} {Variation} and {Blood} {Pressure} {Responses}.
    \newblock {\em Diabetic Medicine}, 9(2):166--175, 1992.
    
    \end{thebibliography}
      
}

\end{document}